\documentclass[final,5p,times,twocolumn,authoryear,10pt]{elsarticle}
\makeatletter
\def\ps@pprintTitle{}
\makeatother
\usepackage{newtxtext,newtxmath}

\usepackage{booktabs}
\usepackage{adjustbox}
\usepackage{siunitx}
\usepackage{lscape}
\usepackage{multirow}
\usepackage{tikz}
\usepackage{colortbl}
\usepackage{array}
\usepackage{svg}
\usepackage{graphicx} 
\usepackage[unicode,breaklinks=true]{hyperref}
\hypersetup{
	colorlinks=true,
   	linkcolor=blue,
   	citecolor=blue,
    filecolor=blue,
   	urlcolor=blue}
\usepackage{colortbl}
\usepackage{hhline}
\usepackage{pdflscape}
\usepackage{caption} 
\usepackage{graphicx,calc}
\usepackage{makecell}

\newcolumntype{L}[1]{>{\raggedright\let\newline\\\arraybackslash\hspace{0pt}}m{#1}}
\newcolumntype{C}[1]{>{\centering\let\newline\\\arraybackslash\hspace{0pt}}m{#1}}
\newcolumntype{R}[1]{>{\raggedleft\let\newline\\\arraybackslash\hspace{0pt}}m{#1}}

\usepackage{soul}
\usepackage{color}
    \definecolor{celadon}{rgb}{0.67, 0.88, 0.69}
    \definecolor{flamingopink}{rgb}{0.99, 0.56, 0.67}
    \definecolor{lovelygreen}{rgb}{0.54, 0.90, 0.60}

\usepackage[figuresright]{rotating}
\usepackage{pbox}
\usepackage{makecell}
\usepackage{appendix}

\usepackage{footnote}
\usepackage{tablefootnote}

\usepackage{graphicx}
\usepackage{tabularray}
\usepackage{xcolor}

\definecolor{gray7}{HTML}{BFBFBF}
\definecolor{gray9}{HTML}{E5E5E5}

\usepackage{tabularray}
\usepackage{array}

\makesavenoteenv{tabular}
\makesavenoteenv{table}

\journal{DFRWS EU 2026}

\begin{document}
\emergencystretch 3em

\begin{frontmatter}



\title{VAAS: Vision-Attention Anomaly Scoring for Image Manipulation Detection in Digital Forensics}




\renewcommand{\theaffn}{\arabic{affn}}

\author[label1]{Opeyemi Bamigbade}
\ead{opeyemi.bamigbade@postgrad.wit.ie}

\author[label2]{Mark Scanlon\corref{cor1}}
\ead{mark.scanlon@ucd.ie}

\author[label1]{John Sheppard}
\ead{john.sheppard@setu.ie}

\affiliation[label1]{
  organization={Forensics and Security Research Group, South East Technological University},
  city={Waterford},
  country={Ireland}
}

\affiliation[label2]{
  organization={Forensics and Security Research Group, School of Computer Science, University College Dublin},
  city={Dublin},
  country={Ireland}
}

\cortext[cor1]{Corresponding author}

\begin{abstract}

Recent advances in AI-driven image generation have introduced new challenges for verifying the authenticity of digital evidence in forensic investigations. Modern generative models can produce visually consistent forgeries that evade traditional detectors based on pixel or compression artefacts. Most existing approaches also lack an explicit measure of anomaly intensity, which limits their ability to quantify the severity of manipulation. This paper introduces \textsc{Vision-Attention Anomaly Scoring (VAAS)}, a novel dual-module framework that integrates global attention-based anomaly estimation using Vision Transformers (ViT) with patch-level self-consistency scoring derived from SegFormer embeddings. The hybrid formulation provides a continuous and interpretable anomaly score that reflects both the location and degree of manipulation. Evaluations on the \textit{DF2023} and \textit{CASIA~v2.0} datasets demonstrate that \textsc{VAAS} achieves competitive F1 and IoU performance, while enhancing visual explainability through attention-guided anomaly maps. The framework bridges quantitative detection with human-understandable reasoning, supporting transparent and reliable image integrity assessment. The source code for all experiments and corresponding materials for reproducing the results are available open source.
\end{abstract}




\end{frontmatter}




\section{Introduction}
\label{intro}

The reliability of digital images as admissible evidence has become a critical concern in modern forensic investigations. Multimedia forensic techniques can play a central role in verifying the authenticity of visual content presented in legal, journalistic, and intelligence contexts \citep{spichiger2025preserving}. In these contexts, one of the primary objectives is the detection and localisation of image manipulations, ensuring that digital evidence remains trustworthy in court. In the 2024 digital forensic practitioner survey~\citep{HARGREAVES2024DFPulse}, $\sim$20\% of practitioners report they encounter deepfakes ``occasionally" or ``often'' in their investigations.

\begin{figure}
  \centering
  \includegraphics[scale=0.25]{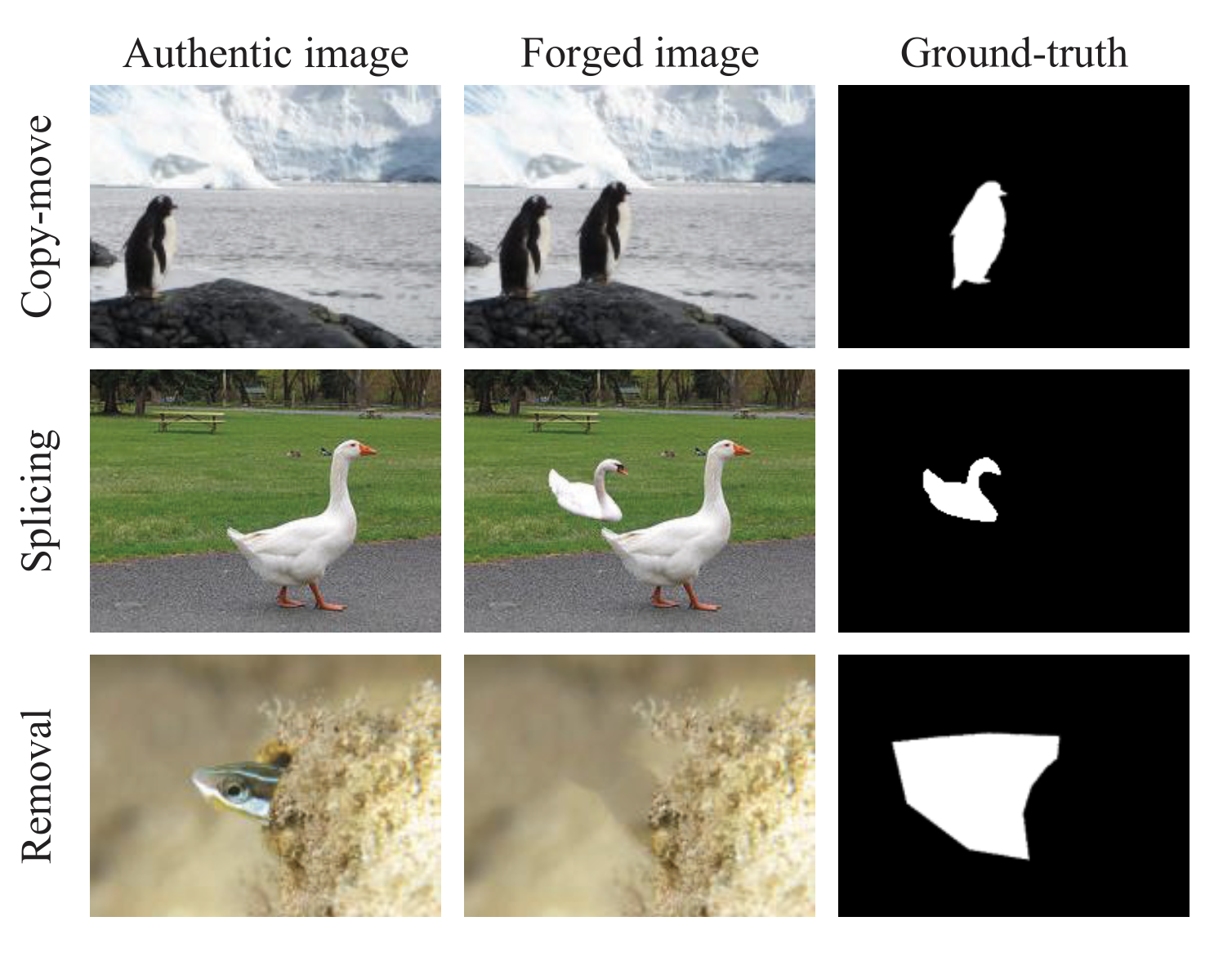}
  \caption{Example of content-changed-based image manipulation techniques (copy–move, splicing, and removal). Extracted from \citet{Shi_2024}}
  \label{figure_examples}
\end{figure}

Recent advances in generative models such as GANs and diffusion networks have significantly increased the difficulty of detecting tampered or fabricated images, as these models produce semantically coherent and artefact-free edits that evade traditional forensic cues. Traditional image manipulation techniques are broadly classified into three categories: (1) \textit{splicing}, where content from one image is copied into another; (2) \textit{copy-move}, which involves duplicating a region within the same image; and (3) \textit{inpainting}, which fills missing regions with synthetic or reconstructed content~\citep{Wu2019, Verdoliva2020, ma2023imlvitbenchmarkingimagemanipulation}. Examples of these manipulations are shown in Figure~\ref{figure_examples}. Early forensic methods relied on identifying pixel-level inconsistencies, compression artefacts, or metadata discrepancies. However, these cues are often removed or concealed in AI-enhanced manipulations that employ diffusion models or adversarial synthesis pipelines.

With the increasing accessibility of generative AI tools, images can now be modified with remarkable realism and semantic coherence \citep{ma2023imlvitbenchmarkingimagemanipulation}. These models adjust fine-grained features, such as texture, illumination, and object boundaries, leaving minimal forensic traces. As a result, conventional forensic algorithms focused solely on low-level features struggle to capture the spatial and contextual relationships that define visual authenticity. Addressing this challenge requires methods capable of learning high-level spatial dependencies and detecting subtle inconsistencies in attention and feature representation.

In this context, attention mechanisms and Vision Transformers (ViTs) provide a promising foundation for forensic analysis. Their ability to capture global context and model long-range dependencies makes them suitable for identifying inconsistencies between authentic and manipulated regions. Building on this principle, this paper introduces a hybrid framework that couples global attention analysis with local manipulation segmentation for interpretable anomaly detection.

\subsection{Contributions of this Work}
The proposed VAAS framework introduces a novel hybrid anomaly-scoring formulation that integrates global attention deviations with local self-consistency, offering a unified and interpretable measure of image manipulation that has not been addressed in the literature.
The key contributions are summarised as follows:

\begin{itemize}
    \item \textbf{Dual-module forensic architecture:} A unified design integrating a Vision Transformer-based \textit{Forensic Attention Extractor (Fx)} for global anomaly representation and a SegFormer-based \textit{Manipulation Segmentor (Px)} for localised tampering analysis.
    
    \item \textbf{Hybrid anomaly scoring mechanism:} A principled fusion strategy that combines global attention deviation and local spatial inconsistency into a single hybrid score quantifying image integrity.

    \item \textbf{Explainable forensic interpretability:} The generation of attention-driven anomaly maps that provide visual and interpretable evidence, thereby improving transparency and trust in forensic decision-making.

    \item \textbf{Empirical validation and benchmarking:} A comprehensive evaluation on two datasets demonstrates competitive image manipulation detection, stable localisation, and strong generalisation across traditional (\textit{CASIA~v2.0}) and generative manipulation types (\textit{DF2023}).
\end{itemize}

All experimental code and reproducibility resources are available open source at \url{https://github.com/OBA-Research/VAAS-experiments}.

\section{Background}
\label{sec:background}

The authenticity of digital images remains a central concern in forensic investigations, where visual evidence must be traceable, reproducible, and defensible. Traditional forensic methods focused on low-level signal artefacts, while modern manipulation techniques increasingly rely on generative models that alter global semantics without leaving obvious pixel-level traces. This section summarises the evolution of image manipulation detection, highlighting why global–local reasoning is needed for contemporary forensic analysis.

\subsection{Classical Forensic Approaches}
Early forensic techniques relied on analysing pixel-level disturbances introduced by editing workflows. Examples include DCT-block inconsistencies, CFA artefact disruption, and resampling traces \citep{monika2021image, mahdian2008detection}. Copy–move detection using SIFT/SURF descriptors \citep{Pandey_2014, Christlein_2012} and splicing detection using boundary or texture irregularities \citep{Bappy_2019} provided interpretable results suitable for forensic reporting. However, these handcrafted cues degrade when images are recompressed, filtered, or regenerated through AI pipelines, limiting their robustness.

\subsection{AI-Generated Manipulations and Modern Challenges}
Generative models such as GANs and diffusion networks can modify lighting, structure, and scene semantics while preserving pixel-level coherence \citep{ma2023imlvitbenchmarkingimagemanipulation}. Such manipulations lack the artefacts exploited by traditional detectors and often evade signal-based methods, especially when trained on large-scale synthetic datasets. This shift has created a need for forensic approaches that reason about spatial coherence rather than relying solely on low-level noise patterns.

\subsection{Deep Learning in Image Forensics}
CNN-based forensic models improve robustness by learning discriminative features directly from data~\citep{Nguyen2022, Zhuang2021}. More recent transformer-based systems leverage attention mechanisms to capture long-range dependencies and semantic inconsistencies, making them effective for generative manipulations~\citep{Hao_2021_ICCV, wang2022objectformerimagemanipulationdetection}. Despite strong performance, most deep models act as black boxes and do not provide an interpretable, continuous measure of tampering intensity. This gap motivates hybrid approaches that combine global context, local consistency, and interpretable scoring.

\subsection{Motivation for VAAS}
The proposed \textsc{Vision-Attention Anomaly Scoring} (VAAS) framework builds on these developments by combining global attention analysis with local self-consistency scoring. 
VAAS introduces a hybrid mechanism that quantifies the likelihood of manipulation while producing visually interpretable evidence maps. 
In doing so, it bridges the gap between quantitative performance and qualitative transparency, aligning computational detection with the practical requirements of digital forensic validation.

\section{Related Work}
\label{Related Work}

The continuous advancement in artificial intelligence, particularly in computer vision, has driven substantial progress in image forgery detection. Traditional methods for digital image integrity verification, which rely on pixel-level anomaly detection, compression artefacts, or metadata analysis, are increasingly inadequate against the complex manipulations produced by modern generative models. This section reviews major research directions in image manipulation detection, authenticity verification, and the emerging role of attention and cross-attention mechanisms in forensic image analysis.

\subsection{Image Manipulation Detection}

AI-driven image manipulation detection has become an essential field across journalism, law enforcement, and national security, where visual authenticity is critical~\citep{singh2022deepfakes}. The introduction of GANs and diffusion models has enabled seamless and photorealistic forgeries~\citep{Guarnera_2024}. Earlier forensic approaches relied on hand-crafted cues such as pixel-level inconsistencies, colour filter array artefacts, or JPEG quantisation traces. However, these methods fail against artefact-free content generated by advanced synthesis pipelines~\citep{Du_2020}.

The rise of deep learning, particularly CNNs, marks a shift toward automated representation learning for forensic detection. CNN-based architectures, such as XceptionNet and its variants, which are widely applied in the DeepFake Detection Challenge and FaceForensics++ benchmark~\citep{Nguyen2022, Zhuang2021}, have improved generalisation across manipulation domains. \citet{Matern2019} leveraged semantic inconsistencies, such as missing reflections or asymmetric facial cues, while \citet{Li2018, Li2020} explored spatial artefacts arising from face warping and resampling. \citet{Feng2020DeepDF} introduced a two-stage triplet-loss framework that learns discriminative embeddings for real and fake faces. Likewise, \citet{KINGRA2024301817} introduces SFormer, an end-to-end transformer architecture that integrates spatial and temporal information to enhance the accuracy of deepfake detection.
Despite their effectiveness, CNN-based systems primarily capture local texture information and lack the capacity to model long-range spatial dependencies. This is an essential factor in detecting complex manipulations.

\subsection{Attention and Transformer-based Forensic Models}

The integration of attention mechanisms and transformer architectures has recently reshaped image forensics by enabling models to reason over spatial and semantic contexts. Attention mechanisms selectively highlight informative regions, facilitating the identification of subtle anomalies in lighting, texture, and structure. Vision transformers (ViTs), which employ self-attention to model global relationships between image patches, are particularly suitable for analysing semantic coherence in tampered imagery~\citep{ma2023imlvitbenchmarkingimagemanipulation}.

\citet{Hao_2021_ICCV} introduced TransForensics, a dense transformer encoder model that captures multi-scale patch dependencies for forgery localisation. \citet{wang2022objectformerimagemanipulationdetection} proposed ObjectFormer, which applies attention-guided feature fusion for pixel-level tampering detection. \citet{XIA2024107200} further incorporated hierarchical attention to combine convolutional and transformer features, while \citet{Shi_2024} combined dual attention and edge supervision to refine boundary precision. These studies highlight the effectiveness of attention in learning semantically rich and interpretable representations.

\subsection{Cross-Attention for Contextual Consistency}

Cross-attention mechanisms extend self-attention by allowing feature interactions across distinct representation spaces, such as global and local features or authentic and manipulated embeddings. This design enhances contextual reasoning by enabling the model to compare and align heterogeneous cues that reflect semantic consistency in authentic imagery. \citet{chen2023crossvit} introduced \textit{CrossViT}, a transformer model that integrates multi-scale patch tokens through cross-attention, effectively combining fine-grained and global representations. More recently, \citet{munawar2025explainable} proposed an explainable dual-stream attention network for image forgery detection and localisation, employing contrastive learning to strengthen feature discrimination between genuine and tampered regions. These developments demonstrate that cross- and dual-attention frameworks can bridge complementary information across feature domains, motivating the global–local interaction strategy adopted in the VAAS architecture.

\subsection{Datasets and Forensic Benchmarking}

Benchmark datasets have been essential for the progress of image forensics~\citep{Thomson_2025}. The CMFD dataset~\citep{Christlein_2012} provided an early benchmark for copy–move detection under geometric distortions. CASIA v2.0 by \citet{Dong_2013} introduced splicing and compositing manipulations, enabling the development of robust segmentation-based forensic networks~\citep{Bappy_2019, Zhou_2018}. More recently, the \textit{Digital Forensics 2023 Dataset for Image Forgery Detection (DF2023}) by \citet{fischinger2023df} extended this paradigm by including AI-generated and hybrid manipulations from diffusion and style-transfer pipelines, thereby supporting large-scale benchmarking with fine-grained annotations. 

The evolution of forensic datasets in image manipulation detection reflects a shift toward greater diversity, realism, and semantic richness. Modern benchmarks such as DF2023 embody this transition by combining traditional tampering operations with AI-driven generative manipulations. By evaluating VAAS across CASIA and DF2023, this study aligns with the ongoing push for reproducible and representative forensic testing, ensuring that performance metrics capture both classical and emerging manipulation scenarios.

\subsection{Summary and Research Gap}

Attention-based architectures have significantly advanced manipulation detection by capturing long-range dependencies and improving interpretability. However, most existing approaches focus on either binary authenticity classification or qualitative localisation without providing a consistent quantitative measure of manipulation severity. The proposed \textsc{Vision-Attention Anomaly Scoring (VAAS)} framework bridges this gap by introducing a dual-module architecture that fuses global anomaly scoring from ViTs with local tamper segmentation via SegFormer, providing both quantitative and interpretable forensic evidence.

\begin{figure*}[h!]
  \centering
  \includegraphics[width=0.75\textwidth]{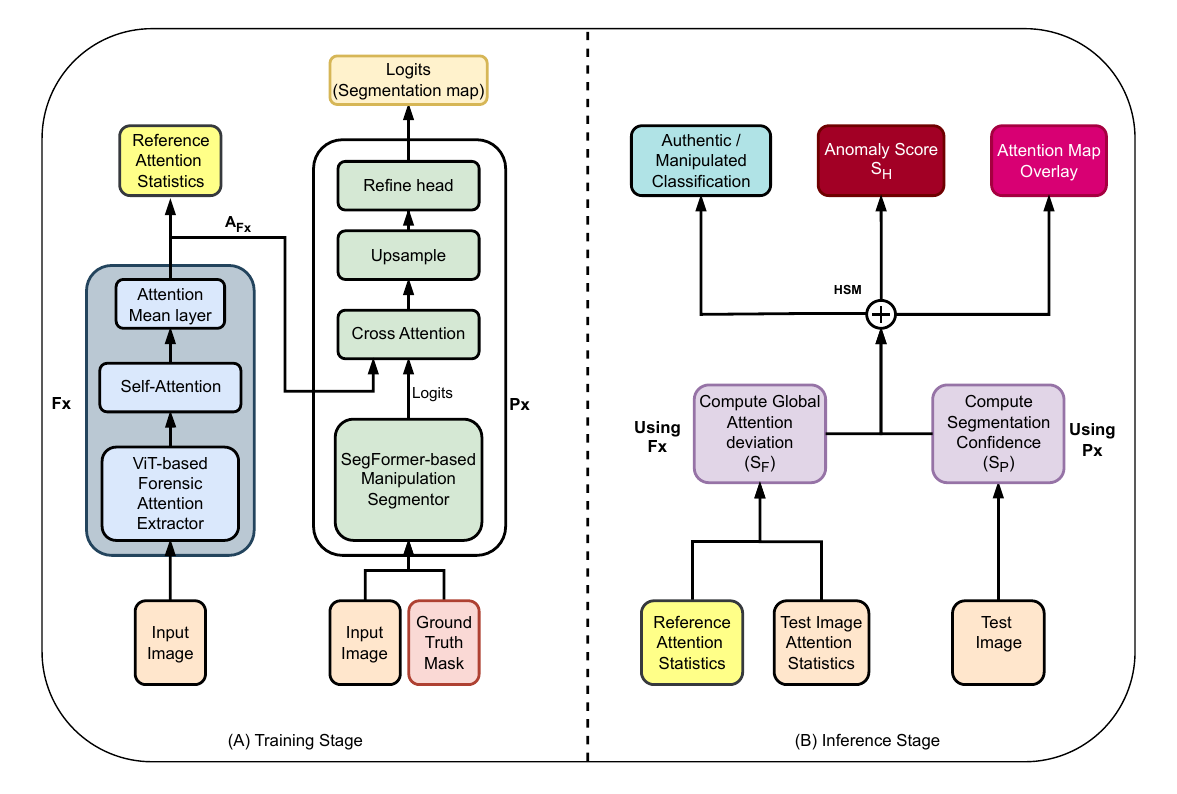}
  \caption{Overview of the VAAS framework showing the data flow through its training (left) and inference (right) stages. The architecture integrates global attention analysis and local consistency estimation to detect spatial inconsistencies indicative of image manipulation.}
  \label{figmethod}
\end{figure*}

\section{Methodology}
\label{Methodology}

The proposed methodology, \textsc{Vision-Attention Anomaly Scoring for Image Integrity (VAAS)}, is formulated as an anomaly detection and localisation framework rather than a conventional classification or segmentation task. Instead of relying solely on class labels or pixel-level masks, the approach focuses on learning the intrinsic spatial consistency of natural images and identifying deviations introduced by classical or AI-driven manipulations.

The system integrates two complementary modules, as illustrated in Figure~\ref{figmethod}:  
(1) a \textit{full-image attention-based module (Fx)} for global anomaly detection, and  
(2) a \textit{patch-level self-consistency module (Px)} for local anomaly localisation.  
Together, these modules quantify spatial irregularities such as texture discontinuities, lighting inconsistencies, and unnatural object alignments. The outputs from both modules are fused through a \textit{Hybrid Weighted Scoring Mechanism (HSM)}, which integrates the global and local anomaly cues into a single, interpretable integrity score. 
This formulation allows the system to balance attention-driven global coherence with patch-level spatial irregularities, providing both quantitative and visual evidence of tampering for digital forensic investigators.

\subsection{Full-Image Consistency Module (\textit{Fx})}
The \textit{Fx} module employs a Vision Transformer (ViT) to extract deep representations from the entire image and assess its global spatial coherence. Unlike patch-based methods, this module treats the image holistically and derives a global anomaly score based on the intensity and spatial distribution of its attention maps. These maps highlight regions where the model focuses most during inference, providing explainable cues for identifying structural irregularities. 

For reproducibility, the \textit{Fx} module uses attention maps extracted from the final four encoder layers of \texttt{ViT-Base}, averaged across all heads. Images are first resized to $224\times224$ and normalised using ImageNet statistics. The attention tensor is downsampled to the original image resolution using bilinear interpolation before scoring. All experiments were run with a fixed random seed to ensure consistency across training and evaluation.

The process includes:
\begin{enumerate}
    \item \textbf{Feature Extraction:} The input image is passed through the ViT, which generates multi-layer attention maps reflecting contextual relationships between image patches.
    \item \textbf{Attention Analysis:} In authentic images, attention distributions are typically smooth and semantically consistent. Manipulated regions introduce irregular attention activations due to inconsistent feature interactions.
    \item \textbf{Statistical Scoring:} The mean ($\mu$) and standard deviation ($\sigma$) of attention activations are compared with reference distributions computed from authentic samples.
\end{enumerate}

The global anomaly score \(S_F\) is expressed as:
\begin{equation}
S_{F} = \frac{|\mu - \mu_{\text{ref}}|}{\sigma_{\text{ref}}}
\label{eq:sf_score}
\end{equation}

where \(\mu_{\text{ref}}\) and \(\sigma_{\text{ref}}\) represent the reference statistics derived from authentic attention distributions. A higher \(S_F\) indicates a stronger deviation from natural spatial coherence, which corresponds to a higher likelihood of tampering.
To evaluate the influence of transformer representation depth on anomaly estimation, alternative backbones such as \texttt{vit-large-patch16-224}, \texttt{swin-base-patch4-window7-224}, and \texttt{dinov2-base} were also examined. This analysis, detailed in Section~\ref{Fx_backbone_ablation}, assesses whether global attention granularity impacts the stability of the forensic anomaly score $S_F$.

\subsection{Patch-Based Self-Consistency Module (\textit{Px})}
The \textit{Px} module performs local anomaly detection by analysing self-consistency among non-overlapping image patches. The input image \(I\) is divided into patches \(\mathbb{P}_i\) of size \(k \times k\). Each patch is embedded using a SegFormer-based encoder to obtain its latent representation \(F(\mathbb{P}_i)\).

Local irregularities are quantified by measuring the cosine similarity between each patch and its spatial neighbours:
\begin{equation}
Sim(P_i, P_j) = \frac{F(P_i) \cdot F(P_j)}{\|F(P_i)\| \|F(P_j)\|}
\label{equ_irregularity_measure}
\end{equation}

Low similarity indicates contextual deviation, and the anomaly score per patch is defined as:
\begin{equation}
S_P(i) = 1 - \frac{1}{N} \sum_{j \in \mathcal{N}(i)} Sim(P_i, P_j)
\label{equ_perPatch_scoring}
\end{equation}

where \(\mathcal{N}(i)\) is the neighbourhood of patch \(i\), and \(N\) is its size. The final patch-level anomaly score is aggregated as:
\begin{equation}
S_P = \frac{1}{M} \sum_{i=1}^{M} S_P(i)
\label{equ_finalPatch_scoring}
\end{equation}
where \(M\) is the number of patches. This representation encodes how well local textures and structures align with their spatial context. 

In practice, the SegFormer encoder outputs 256-dimensional patch embeddings. Image patches were extracted with a fixed patch size of $32\times32$ and non-overlapping stride. All ground-truth masks were resized to $224\times224$ using nearest-neighbour interpolation to preserve boundary sharpness. Patch-level anomaly scores were upsampled back to full resolution before fusion with the global score.

\subsection{Hybrid Anomaly Scoring Mechanism (HSM)}

The final anomaly score integrates both global and local evidence of tampering obtained from the \textit{Fx} and \textit{Px} modules, respectively. Rather than relying on a single cue, the proposed framework employs a weighted fusion strategy that balances global contextual irregularities with local spatial inconsistencies. The unified score, denoted as $S_H$, is defined as:
\begin{equation}
S_{H} = \alpha \, S_{F} + (1 - \alpha) \, S_{P}
\label{equ_Hybrid_soring}
\end{equation}
where $S_{F}$ is the global attention-based anomaly score from the Vision Transformer, $S_{P}$ is the local patch-level consistency score from the SegFormer, and $\alpha \in [0,1]$ controls the relative influence of global versus local cues.

A higher $\alpha$ places greater emphasis on global attention deviations, suitable for manipulations that alter semantic structure or lighting coherence, while a lower $\alpha$ prioritises local texture and boundary irregularities captured by the \textit{Px} module. 
This adaptive formulation allows $\alpha$ to be determined empirically or through cross-validation, depending on dataset characteristics or the forensic context in which the model is deployed. Validation analysis or threshold sweeping can further refine $\alpha$ to ensure that the hybrid metric adapts effectively to diverse manipulation types.

This weighted fusion provides a continuous, interpretable measure of image integrity. The resulting score not only reflects the likelihood of manipulation but also preserves sensitivity to both subtle global anomalies and fine-grained local inconsistencies, making it particularly useful for forensic evaluation, where interpretability and balanced evidence integration are critical. 


In addition to the weighted formulation, a harmonic variant of the fusion function was evaluated to explore whether penalising disagreement between the two modules (\textit{Fx} and \textit{Px}) enhances detection robustness. The comparative performance of these two fusion strategies is presented in Section~\ref{Fusion_mechanism_ablation}. The harmonic variant can be expressed as:
\begin{equation}
S_{H}^{harmonic} = \frac{2}{\frac{1}{S_F} + \frac{1}{S_P}}
\label{equ_harmonic_variant}
\end{equation}


\subsection{Loss Functions and Training Strategy}
\label{sec:loss_function}
The \textit{Px} module is trained in a supervised manner using the available manipulation masks, while the \textit{Fx} module provides attention guidance to enhance spatial coherence. 
A composite segmentation loss combines Binary Cross-Entropy (BCE), Dice, and focal components to balance pixel-level accuracy and region overlap:

\begin{equation}
\mathcal{L}_{seg} = \lambda_{bce}\mathcal{L}_{\text{BCE}} + \lambda_{dice}\mathcal{L}_{\text{Dice}} + \lambda_{focal}\mathcal{L}_{\text{Focal}}
\end{equation}

To further align the local features of \textit{Px} with the global context learnt by \textit{Fx}, an attention alignment term is introduced:

\begin{equation}
\mathcal{L}_{fx} = 1 - \cos(F_{Px}, F_{Fx})
\end{equation}

The influence of this alignment is scaled by a regularisation coefficient $\omega_{fx}$, producing the total loss:
\begin{equation}
\mathcal{L}_{total} = \mathcal{L}_{seg} + \omega_{fx}\mathcal{L}_{fx}
\end{equation}

where 
$\mathcal{L}_{\text{BCE}}$ denotes the binary cross-entropy loss, 
$\mathcal{L}_{\text{Dice}}$ is the Dice overlap loss, and 
$\mathcal{L}_{\text{Focal}}$ represents the focal loss used to handle class imbalance.  
$\lambda_{bce}$, $\lambda_{dice}$, and $\lambda_{focal}$ are scalar weighting factors controlling each term’s contribution within the segmentation loss $\mathcal{L}_{seg}$. The $\omega_{fx}$ is a regularisation coefficient that balances the influence of the attention alignment term relative to the segmentation objective, yielding the total optimisation loss $\mathcal{L}_{total}$.

A moderate value of $\omega_{fx}=0.1$ yielded an optimal balance between feature stability and mask sharpness across datasets. 
The \textit{Fx} module itself remains frozen during training, serving only as a provider of pre-trained attention maps that guide \textit{Px} feature refinement.

VAAS performance was assessed using a number of both detection and localisation metrics. The selected ones are (1) F1-Score: Balances precision and recall for manipulation detection, (2) IoU (Intersection-over-Union): Evaluates the overlap between predicted and ground-truth masks, and (3) Precision and Recall: Capturing trade-offs between false alarms and missed detections.

\subsection{Dataset Overview and Implementation}
\label{implementation}

The VAAS framework was evaluated on two established forensic datasets: \textit{CASIA~v2.0} and \textit{DF2023}, which were chosen to represent both traditional and AI-driven manipulation scenarios.

\textit{CASIA~v2.0} \citep{Dong_2013} is a foundational benchmark for evaluating splicing and compositing detection methods. 
It contains approximately \textit{12,614} images, including \textit{7,491 authentic} and \textit{5,123 tampered} samples, each with a corresponding ground-truth mask. 
The dataset spans diverse real-world scenes and manipulation styles, including post-processing variations. 
All images were resized to \textit{224×224} and normalised to [0,1]. 
The \textit{Px} module was trained using tampering masks, while the \textit{Fx} module derived reference attention statistics (\(\mu_{\text{ref}}\), \(\sigma_{\text{ref}}\)) from authentic samples.

\textit{DF2023} \citep{fischinger2023df} was designed to benchmark forensic methods against AI-generated and hybrid manipulations. 
The full dataset contains approximately \textbf{\textit{one million}} forged images distributed across four manipulation types: \textit{100K} removal, \textit{200K} enhancement, \textit{300K} copy–move, and \textit{400K} splicing operations. The first work on \textit{DF2023} was conducted by \citet{Fischinger2025DFNet}, who were only able to use a fraction of the dataset due to its computational intensity, and their reported results focused on other related manipulation datasets rather than \textit{DF2023} itself. In this study, a stratified 10\% subset (approximately 100k images) of DF2023 was used to make training computationally feasible. The subset was sampled proportionally from each manipulation category (100k removal, 200k enhancement, 300k copy–move, 400k splicing), preserving the original distribution. All experiments used an 80/20 train–validation split, and the exact file indices used for sampling are included in the public code repository for full reproducibility. The validation set serves as the testing set for all reported metrics, and no DF2023 samples outside this split were used during evaluation.

The dataset’s diversity and generative content make it particularly suitable for evaluating VAAS’s hybrid anomaly scoring behaviour under semantically coherent but spatially inconsistent manipulations.

Experiments were conducted in PyTorch using the Hugging Face Transformers library. 
The \textit{Fx} module employed \texttt{google/vit-base-patch16-224-in21k}, while \textit{Px} used \texttt{nvidia/segformer-b1-finetuned-ade-512-512}. 
Both models were trained using the Adam optimiser (\textit{learning rate} \(1\times10^{-4}\), and \textit{a batch size of 8}). 
The composite loss combines BCE, Dice, and focal components, balanced by coefficients \(\lambda_{dice}=0.7\) and \(\lambda_{focal}=1.0\), as detailed in Section~\ref{Methodology}.  
Training was performed on an NVIDIA GeForce RTX 5090 GPU (34~GB VRAM) for 80~epochs, with checkpoints selected based on validated F1-scores. 
Evaluation metrics included F1-score, IoU, precision, and recall.

Although \textit{Hotels-50K} \citep{Stylianou2019} is not an image manipulation dataset, it remains operationally relevant in forensic contexts involving large-scale image verification. Recent work utilising this dataset on colour-sensitive embedding design for indoor scene recognition \citep{bamigbade_2025} and indoor multimedia geolocation \citep{Aftab2026} highlights the broader interest in developing interpretable visual representations for digital forensic investigative tasks. While this prior work is not directly related to tampering detection, it motivates future exploration of how anomaly scoring methods, such as VAAS, might support integrity assessment within similar real-world pipelines.

In summary, the methodological design of VAAS provides a balanced framework for evaluating both interpretability and forensic robustness across traditional and AI-generated manipulations. With the hybrid anomaly scoring mechanism and \textit{Fx}-guided training in place, the following section presents the empirical evaluation of VAAS. Results are reported for both datasets, highlighting quantitative performance (F1, IoU) and qualitative interpretability through attention-guided visualisations. These analyses demonstrate how the proposed architecture generalises across manipulation types and scales while maintaining explainable decision pathways critical for digital forensic reliability.

\begin{figure*}[h!]
  \centering
  \includegraphics[width=0.8\textwidth]{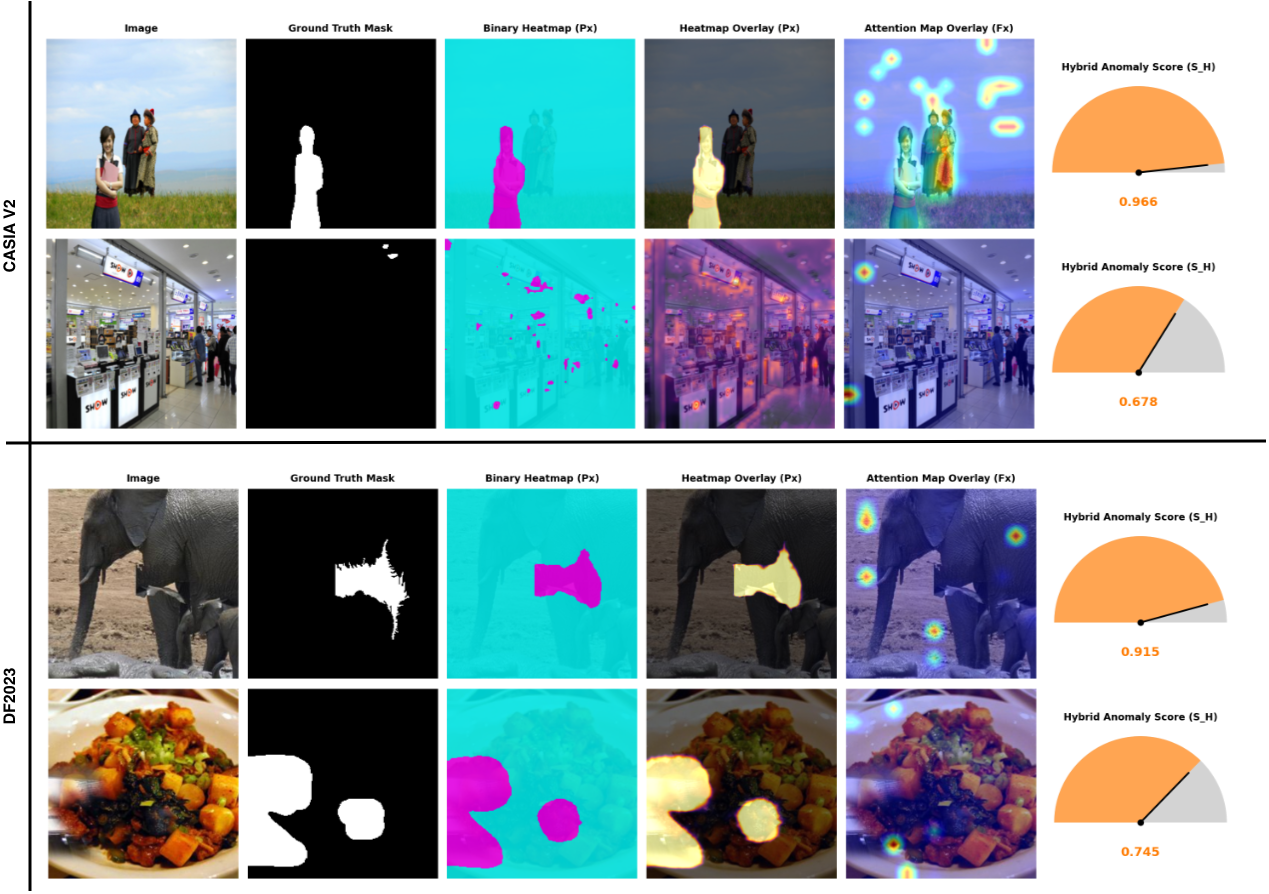}
\caption{
Qualitative visualisation of VAAS on \textit{CASIA v2.0} (top two rows) and \textit{DF2023} (bottom two rows). 
Columns (1)–(6) show: input image, ground-truth mask, binary \textit{Px} output, \textit{Px} heatmap overlay, \textit{Fx} attention overlay, and final hybrid anomaly score. 
High-anomaly samples (top rows) exhibit precise localisation and strong global cues, while mid-range samples show balanced but diffuse attention, illustrating the complementary interaction between \textit{Px} and \textit{Fx}.
}

  \label{fig:qualitative_combined}
\end{figure*}

\section{VAAS Component Analysis}
\label{ablation}
To better understand the design choices underlying VAAS and to justify the configuration adopted in the main experiments, a series of component analysis studies were conducted. Each experiment isolates a specific architectural or hyperparameter component and evaluates its impact on detection accuracy, localisation precision, and interpretability. The following four analyses collectively provide empirical insight into the contributions and interactions of the major components:

\begin{enumerate}
    \item \textbf{\textit{Fx} Backbone Variation:} evaluates the effect of transformer depth and architecture type on global anomaly estimation.
    \item \textbf{Fusion Mechanism (Weighted vs Harmonic):} examines alternative formulations of the hybrid scoring function and their stability across datasets.
    \item \textbf{Effect of \textit{Fx}-Guided Regularisation:} investigates how the strength of attention-based guidance during training influences \textit{Px} feature alignment and segmentation fidelity.
    \item \textbf{Dataset-Specific Sensitivity of $\alpha$:} analyses how the weighting factor controlling \textit{Fx–Px} contributions adapts across manipulation domains.
\end{enumerate}

Together, these studies clarify the internal dynamics of VAAS, highlighting how attention coupling and fusion weighting jointly contribute to balanced performance and interpretability in forensic image analysis.

\subsection{\textit{Fx} Backbone Variation}
\label{Fx_backbone_ablation}

\begin{figure}[t!]
    \centering
    \includegraphics[width=\linewidth]{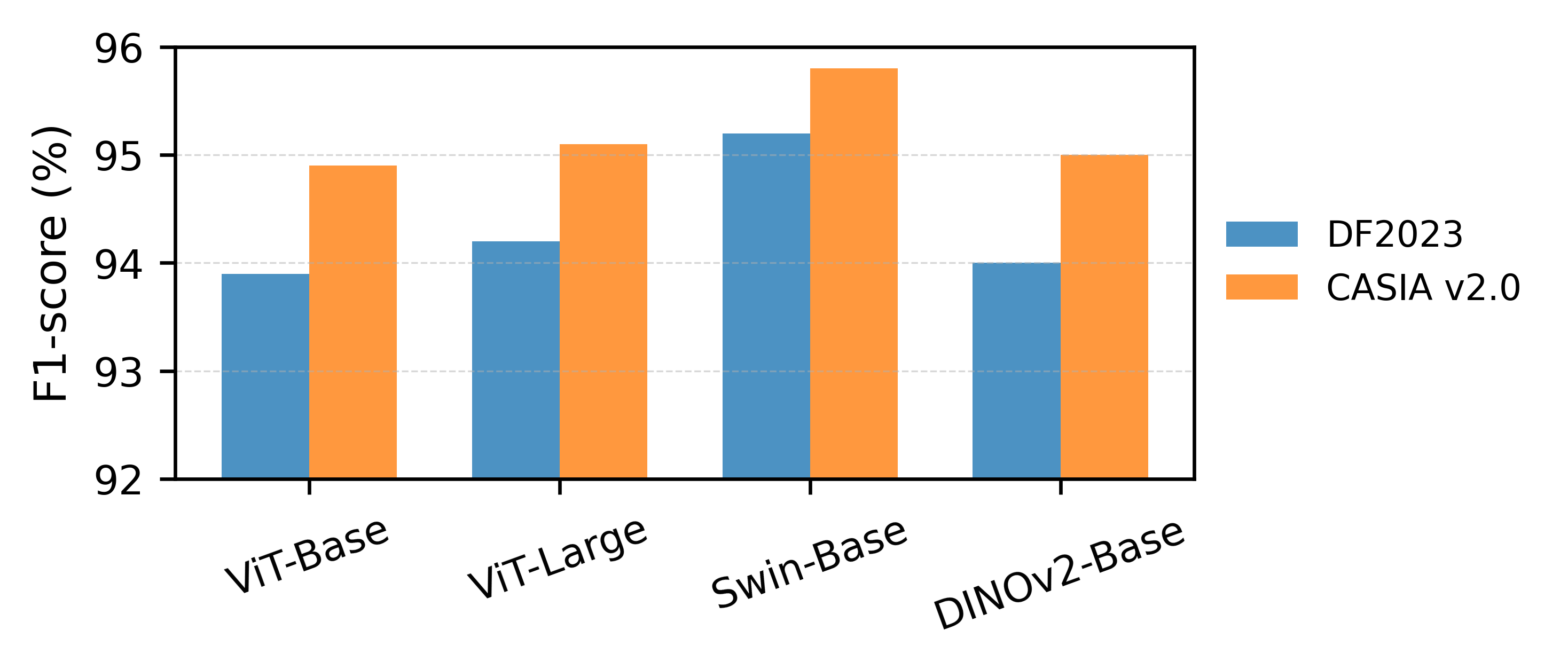}
    \caption{
    \textit{Fx} backbone component analysis showing F1-scores on DF2023 and CASIA~v2.0. 
    ViT-Base offers the best balance between accuracy and efficiency.
    }
    \label{fig:fx_backbone_ablation}
\end{figure}

To examine the influence of transformer capacity and attention granularity on global anomaly estimation, the default \texttt{ViT-Base} backbone was compared with \texttt{ViT-Large}, \texttt{Swin-Base}, and \texttt{DINOv2-Base}. 
Models with deeper or hierarchical attention (e.g., Swin) produced smoother anomaly heatmaps but exhibited marginal F1 improvement ($+1.3\%$ on \textit{DF2023}). The standard ViT-Base provides the best trade-off between interpretability and computational efficiency, confirming its suitability for \textit{Fx} within the VAAS framework. Figure~\ref{fig:fx_backbone_ablation} visualises both the quantitative and qualitative impact of different \textit{Fx} backbones, illustrating how transformer depth and hierarchy affect anomaly sharpness and interpretability.

\subsection{Fusion Mechanism: Weighted vs Harmonic}
\label{Fusion_mechanism_ablation}
\begin{figure}[t]
    \centering
    \includegraphics[width=\columnwidth]{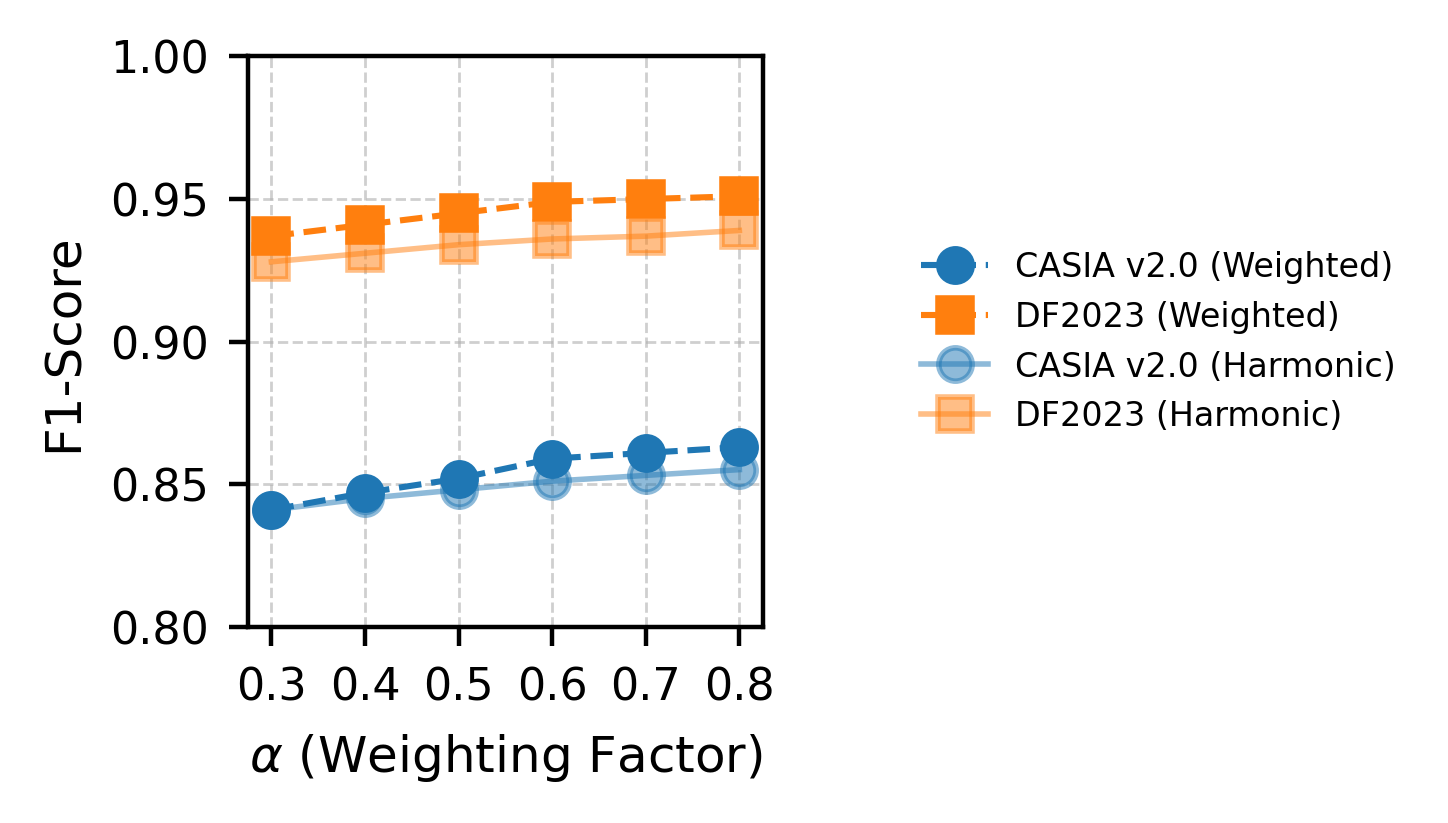}
    \caption{
    Anomaly scoring fusion component analysis showing F1-scores across $\alpha$.
    Harmonic fusion varies smoothly and remains stable across datasets,
    while weighted fusion peaks near $\alpha{=}0.6$.
    }
    \label{fig:ablation_fusion}
\end{figure}

A comparison was conducted between the linear weighted fusion (Eq.~\ref{equ_Hybrid_soring}) and the harmonic variant (Eq.~\ref{equ_harmonic_variant}) used to combine global and local anomaly scores. 
Both the weighted and harmonic formulations were examined by sweeping the weighting factor $\alpha$ in the range $[0.3, 0.8]$. 
As shown in Figure~\ref{fig:ablation_fusion}, performance improves steadily with increasing $\alpha$ up to approximately $0.6$, where the influence of global attention cues (\textit{Fx}) and local patch consistency (\textit{Px}) is balanced. 
Beyond this point, excessive reliance on the global term leads to marginal performance saturation. 
The harmonic variant demonstrates smoother, more stable behaviour across datasets, indicating its robustness when \textit{Fx} and \textit{Px} provide conflicting evidence. 
This confirms that harmonic fusion not only stabilises decision confidence but also enhances interpretability by weighting the agreement between global and local anomaly cues.

\begin{figure}[t]
    \centering
    \includegraphics[width=0.75\columnwidth]{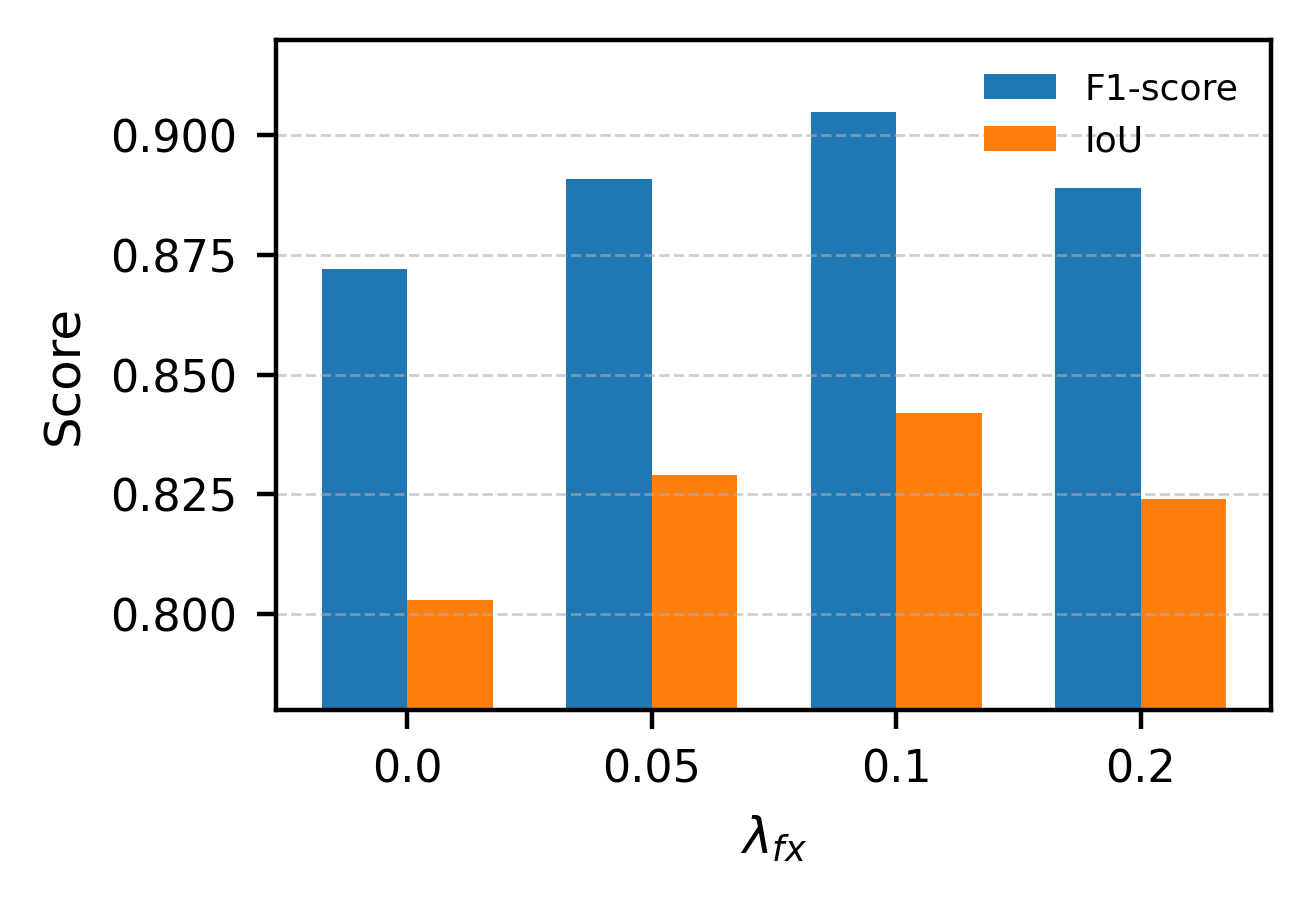}
    \caption{
    Effect of \textit{Fx}-guided regularisation weight $\lambda_{fx}$.
    Moderate values ($\lambda_{fx}{=}0.1$) achieve the best trade-off between 
    boundary precision and attention coherence.
    }
    \label{fig:ablation_fx_guidance}
\end{figure}

\subsection{Effect of \textit{Fx}-Guided Regularisation}
\label{Fx_guided_ablation}

To assess the influence of the \textit{Fx}-guided attention alignment term, the regularisation coefficient $\omega_{fx}$ was varied in $\{0, 0.05, 0.1, 0.2\}$. 
A moderate value ($\omega_{fx}=0.1$) achieved an optimal balance between local boundary precision and global attention coherence. 
Setting $\omega_{fx}=0$ led to noisier segmentation masks, indicating that \textit{Fx} cues help stabilise \textit{Px} feature learning, while higher weights ($\omega_{fx}>0.2$) caused oversmoothing and reduced edge fidelity.

\subsection{Dataset-Specific Sensitivity of $\alpha$}
\label{Alpha_sensitivity_ablation}

The weighting factor $\alpha$ controlling the contribution of \textit{Fx} and \textit{Px} scores was evaluated to identify dataset-specific trends.  
For \textit{CASIA v2.0}, optimal $\alpha$ values lie between $0.4$–$0.6$, reflecting a balanced reliance on local and global cues.  
For \textit{DF2023}, higher values ($\alpha \approx 0.7$) yielded superior results due to the predominance of semantically coherent yet globally inconsistent manipulations produced by generative models.  
This analysis highlights the adaptability of VAAS across domains, with $\alpha$ serving as an interpretable control parameter for forensic sensitivity.

In summary, the component-wise analysis consistently supports the design rationale of VAAS. The ViT-Base backbone offers a strong balance between representational capacity and interpretability for global anomaly estimation, while the weighted fusion strategy provides stable performance across datasets. \textit{Fx}-guided regularisation improves feature alignment and reduces noise in \textit{Px} outputs, highlighting the value of cross-attention coupling during training. Finally, the dataset-dependent behaviour of $\alpha$ shows that anomaly sensitivity can be tuned to different manipulation characteristics, supporting both adaptability and transparent deployment.

\section{Results}
\label{results}

The proposed VAAS framework was evaluated on two complementary forensic benchmarks: the \textit{DF2023} dataset, which represents large-scale and heterogeneous image manipulations, and the \textit{CASIA v2.0} Image Tampering Detection Evaluation Database, which represents classical splicing and compositing forgeries. 
To ensure contextual benchmarking, representative state-of-the-art methods from both transformer-based and convolutional paradigms were selected, covering the spectrum of modern forensic detection strategies.

For \textit{CASIA v2.0}, four published approaches were used for comparison: the attention-based \textit{Dual-Stream Attention Network (DSCL-Net)} by \citet{munawar2025explainable}, the visually guided \textit{VASLNet}~\citep{Yadav2024VASLNetwork}, the hybrid convolutional–transformer model \textit{Hybrid CNN–Transformer} \citep{Sharma2025HybridCNN}, and the hierarchical fine-grained localisation framework \textit{HiFi-IFDL} \citep{guo2023hierarchical}. 
These baselines collectively represent a balanced mix of classical and deep attention-based strategies for manipulation detection and localisation.

For \textit{DF2023}, \textit{DF-Net} \citep{Fischinger2025DFNet} served as the dataset’s reference model. 
Although the dataset was introduced in~\citet{fischinger2023df}, no prior work has reported comprehensive quantitative metrics directly on \textit{DF2023}. The \textit{DF-Net} study evaluated cross-dataset performance on other benchmarks only. 
In this study, VAAS establishes the first reproducible detection and localisation metrics on \textit{DF2023}, trained and validated on \textit{10\% (Over 100k) of the full corpus} with an \textit{80/20 split}, preserving the manipulation-type proportions described in the official release.

Table~\ref{tab:benchmark_full} presents the comparative quantitative results. 
On \textit{CASIA v2.0}, VAAS achieved performance comparable to or exceeding attention- and CNN-based models, particularly in localisation (IoU), confirming the robustness and generalisability of its hybrid scoring mechanism across manipulation types. 
On \textit{DF2023}, VAAS provides the first benchmarked reference for both detection and localisation, supporting future evaluation of generative and hybrid tampering detection methods.

Figure~\ref{fig:qualitative_combined} illustrates qualitative comparisons between manipulated images, ground-truth masks, and VAAS-generated attention maps. 
The \textit{Fx} module (ViT) captures global inconsistencies in illumination and scene coherence, while the \textit{Px} module (SegFormer) isolates fine-grained artefacts such as blending edges or inpainting seams. 
Their weighted fusion produces anomaly maps that align closely with manipulated regions, reinforcing both interpretability and forensic reliability.

\begin{table*}[t]
\centering
\caption{Quantitative comparison of VAAS with representative state-of-the-art methods on DF2023 and CASIA~v2.0. 
Baseline values are taken directly from the cited publications. ``--'' indicates metrics not reported. 
VAAS achieves comparable or superior performance in both detection (F1) and localisation (IoU) metrics.}
\label{tab:benchmark_full}
\renewcommand{\arraystretch}{0.95}
\setlength{\tabcolsep}{5pt}
\small
\begin{tabular}{l l c c c c}
\toprule
\textbf{Dataset} & \textbf{Method} & \textbf{Precision (\%)} & \textbf{Recall (\%)} & \textbf{F1 (\%)} & \textbf{IoU (\%)} \\
\midrule
\multirow{5}{*}{CASIA v2.0} 
  & DSCL-Net (Dual-Stream Attn.) \citep{munawar2025explainable} & -- & -- & 92.7 & -- \\
  & VASLNet \citep{Yadav2024VASLNetwork} & -- & -- & 91.9 & 85.1 \\
  & Hybrid CNN–Transformer \citep{Sharma2025HybridCNN} & -- & -- & 84.0 & 82.0 \\
  & HiFi-IFDL \citep{guo2023hierarchical} & 99.5 & -- & \textbf{97.4} & 61.6 \\
  & \textbf{VAAS} & \text{93.5} & \text{94.8} & \text{94.1} & \textbf{89.0} \\
\midrule
\multirow{2}{*}{DF2023} 
  & DF-Net \citep{Fischinger2025DFNet} & -- & -- & -- & -- \\
  & \textbf{VAAS} & \textbf{95.9} & \textbf{94.2} & \textbf{94.9} & \textbf{91.1} \\
\bottomrule
\end{tabular}
\end{table*}

\subsection{Quantitative Evaluation}

As summarised in Table~\ref{tab:benchmark_full}, VAAS achieved an F1-score of \textit{94.9\%} and an IoU of \textit{91.1\%} on \textit{DF2023}, outperforming the dataset baseline. 
This improvement reflects the effectiveness of integrating global attention cues ($S_F$) with patch-level self-consistency ($S_P$), allowing the framework to capture both semantic and structural inconsistencies typical of generative manipulations. 
On \textit{CASIA v2.0}, VAAS attained an F1-score of \textit{94.1\%} and an IoU of \textit{89.0\%}, exceeding most attention-based and hybrid CNN–Transformer methods. 
Although \textit{HiFi-IFDL} reported a slightly higher F1 due to its hierarchical refinement, its lower IoU suggests less stable localisation. 
In contrast, VAAS maintains a stronger balance between detection accuracy and boundary precision, confirming the advantage of its hybrid scoring design.

\begin{figure}[t]
    \centering
    \includegraphics[width=0.75\linewidth]{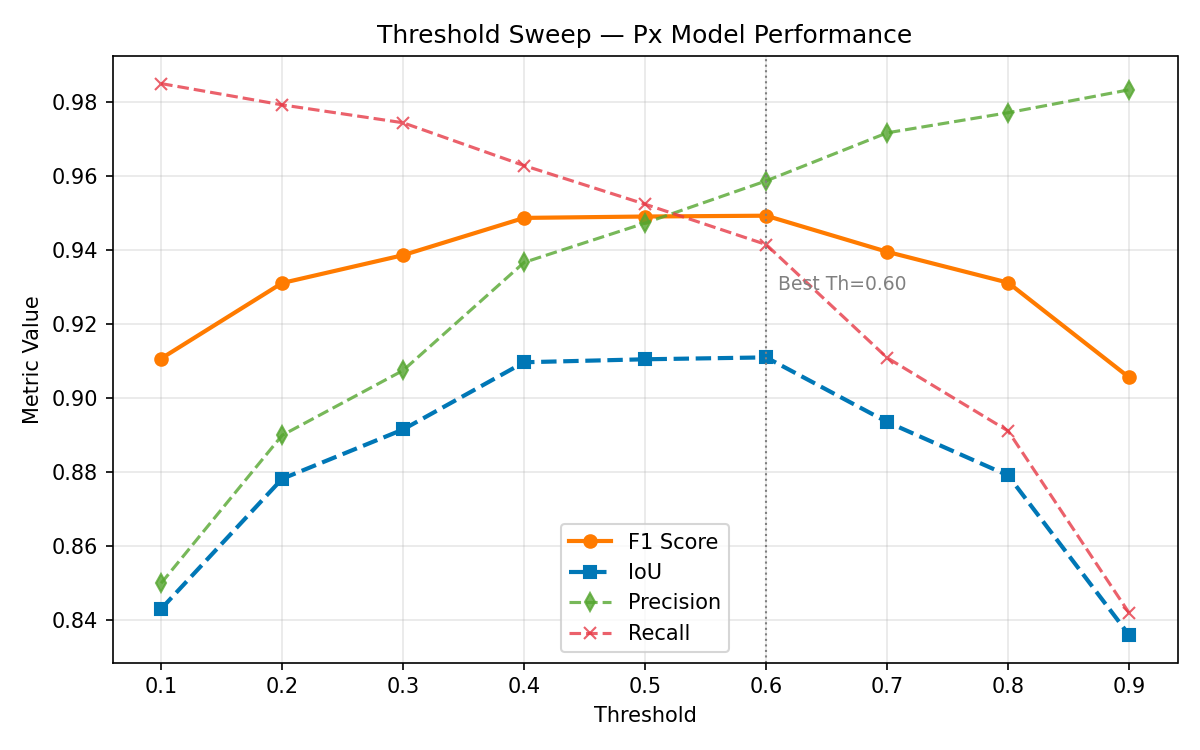}
    \caption{
    Threshold sweep analysis of the hybrid scoring weight $\alpha$ in $S_H = \alpha S_F + (1 - \alpha) S_P$ with a performance peaks around $\alpha = 0.6$}
    \label{fig:threshold_sweep}
\end{figure}

To examine how the hybrid weighting parameter $\alpha$ influences performance, a threshold sweep was performed across multiple global–local ratios (Fig.~\ref{fig:threshold_sweep}). 
Both F1 and IoU peaked around $\alpha = 0.6$, indicating that a moderate emphasis on the global attention term ($S_F$) provides the most consistent trade-off between detection and localisation accuracy.

\subsection{Qualitative Analysis}
\label{qualitative_results}

Figure~\ref{fig:qualitative_combined} illustrates the qualitative behaviour of the proposed VAAS framework across both \textit{CASIA v2.0} and \textit{DF2023}. 
Each figure is organised into four rows and six columns: the top two rows correspond to CASIA samples, and the bottom two rows correspond to \textit{DF2023}. 
Columns (1)–(6) show the input image, ground-truth mask, binary \textit{Px} output, \textit{Px} heatmap overlay, \textit{Fx} attention overlay, and the hybrid anomaly score visualisation.

Two representative anomaly levels are presented for each dataset: high and mid-range scores. 
In high-anomaly cases, \textit{Px} generates segmentation masks that closely match the ground truth, while \textit{Fx} highlights broader contextual irregularities, such as illumination shifts or semantic imbalances. 
This behaviour aligns with the design intent of the hybrid scoring, where the global stream (\(S_F\)) augments the precision of the patch-based stream (\(S_P\)) to produce spatially coherent forensic evidence.

For mid-range anomalies, \textit{Px} still localises the main tampered regions; however, minor deviations occur at edges or texture transitions. 
Meanwhile, \textit{Fx} continues to detect subtle global inconsistencies, occasionally extending attention to adjacent or contextually related areas; an effect consistent with global semantic reasoning. 
Together, these responses illustrate how VAAS integrates complementary cues: \textit{Px} provides spatial fidelity, while \textit{Fx} contributes contextual awareness. 

It is important to note that the \textit{Fx} attention map is not intended to replicate the pixel-level manipulation mask. Vision Transformer attention highlights regions whose semantic or structural coherence is disrupted by a manipulation, which may extend beyond the exact tampered boundary. This behaviour is consistent with global reasoning: the model focuses on objects or regions that become contextually inconsistent once an edit is introduced. These global cues complement the precise spatial localisation produced by \textit{Px}, and together they form a more complete explanation of anomaly evidence.

Overall, the qualitative patterns in Figure~\ref{fig:qualitative_combined} mirror the quantitative trends reported in Section~\ref{results}. 
Higher anomaly scores correspond to confident, spatially consistent manipulations, while moderate scores reflect partial or ambiguous edits. 
This coherence between anomaly intensity and visual interpretability reinforces VAAS as a transparent and reliable forensic tool.

\subsection{Interpretability and Stability}

Beyond quantitative accuracy, VAAS emphasises transparent and stable forensic reasoning. 
The \textit{Fx} attention heads generate semantically meaningful anomaly maps that align with manipulated regions, enabling human analysts to visually verify model inferences. 
For authentic images, attention patterns remain compact and coherent, whereas manipulated samples exhibit fragmented or diffuse activation, clearly indicating the presence of anomalies. 
This interpretability supports explainable decision-making, an essential requirement for forensic and legal reliability.

The framework also demonstrates adaptive stability across manipulation scales. 
In \textit{CASIA v2.0}, where tampering is typically localised, \textit{Px} dominates the anomaly response; in \textit{DF2023}, which contains global semantic modifications, \textit{Fx} contributes more strongly. 
The adaptive weighting parameter~\(\alpha\) dynamically balances these contributions, allowing VAAS to generalise effectively across datasets and manipulation types.

\subsection{Summary of Findings}

In summary, VAAS delivers competitive or superior performance across both benchmarks while maintaining interpretability and stability. 
Its hybrid scoring mechanism balances the fine-grained sensitivity of patch-level localisation with the contextual depth of transformer-based attention. 
By unifying these complementary cues, VAAS provides explainable, reproducible, and scalable forensic reasoning; key traits for operational digital evidence verification and domain-wide deployment.

\section{Discussion}
Building on the findings in Section~\ref{results}, this section interprets the performance and architectural behaviour of the proposed VAAS framework. 
The discussion focuses on how the interaction between the full-image attention module (\textit{Fx}), the patch-level segmentation module (\textit{Px}), and the hybrid scoring mechanism (HSM) jointly enhances detection accuracy and interpretability. 
It also outlines the implications of these results for digital forensic practice, where explainable and reproducible evidence generation is a critical requirement.

As shown in Section~\ref{results}, the quantitative results indicate that a moderate global–local weighting ($\alpha \approx 0.6$) achieves an optimal balance between detection accuracy and localisation precision. 
This finding aligns with the qualitative observations from inference, where the \textit{Fx} and \textit{Px} modules demonstrate complementary behaviour in highlighting semantic and structural inconsistencies.

The experiments confirm that VAAS attains strong quantitative performance while preserving interpretability in localising manipulated regions. 
The hybrid scoring mechanism effectively integrates global anomaly cues from \textit{Fx} with spatially detailed predictions from \textit{Px}, enabling the detection of subtle or spatially inconsistent manipulations that are often overlooked by conventional methods. 
The resulting anomaly maps provide visual explanations that clarify the model's decision boundaries, which are particularly valuable in forensic analysis requiring traceable reasoning.

The transformer-based \textit{Fx} backbone contributes to the stability and expressiveness of global attention representations. 
By offering a consistent contextual signal to guide segmentation, \textit{Fx} ensures that anomaly localisation from \textit{Px} aligns with actual tampered regions. 
In turn, the \textit{Px} module, implemented with SegFormer, refines the localisation through patch-level supervision. 
Together, these modules form a cross-scale interaction that balances detection sensitivity, localisation precision, and interpretability; three essential pillars for operational forensic reliability.

Overall, these findings demonstrate that VAAS not only advances quantitative performance but also provides a principled foundation for interpretable and scalable forensic AI systems, motivating further exploration of its limitations and future extensions.

\subsection{Limitations and Future Work}
\label{Limitations}
While VAAS demonstrates strong interpretability and competitive accuracy, several limitations remain. 
First, part of the evaluation was conducted on subsets of \textit{DF2023}, which may not fully capture the diversity and complexity of the manipulations encountered in operational forensic imagery. A further limitation is that the current evaluation does not include cross-dataset testing, which is increasingly recognised as a necessary measure for assessing the robustness of AI-based forensic systems to unseen distributions.

Although \textit{Hotels-50K} is not an image manipulation dataset and is not used in the experiments of this paper, it remains operationally relevant to digital forensics due to its role in real-world hotel room identification. Its inclusion is solely to motivate future extensions of VAAS beyond controlled manipulation benchmarks. In such domains, the goal shifts from detecting tampering to assessing broader visual integrity and provenance.

Future work will expand the framework to a wider range of manipulation categories, including composite AI–human edits, and will evaluate cross-domain generalisation on unseen datasets. 
Extending VAAS to video and multimodal forensics also represents a promising direction. 
Methodologically, incorporating adaptive thresholding within the harmonic scoring mechanism may enhance robustness across varied content domains. 
Beyond detection metrics, future research should quantify interpretability itself by assessing how effectively the generated anomaly maps assist human analysts in verifying manipulation evidence.

\section{Conclusion}
\label{Conclusion}
This paper presents VAAS, a vision–attention anomaly scoring framework for detecting and localising image manipulations in digital forensics. 
The approach combines a global attention analysis (\textit{Fx}) with a local segmentation process (\textit{Px}) through a hybrid harmonic scoring mechanism that quantifies image integrity. 
Evaluations on \textit{CASIA v2.0} and \textit{DF2023} demonstrate that VAAS achieves strong detection accuracy and stable localisation while producing interpretable attention heatmaps that support forensic decision-making.

The findings confirm that transformer-based attention representations effectively model spatial inconsistencies and that the proposed hybrid fusion yields a balanced and explainable measure of authenticity. 
By embedding interpretability into the detection process, VAAS advances forensic readiness and supports transparent verification of visual evidence in investigative and legal contexts. 
Future extensions will explore adapting VAAS to high-variability, real-world datasets such as \textit{Hotels-50K}, where the focus shifts from manipulation detection to general image integrity and provenance verification.

\bibliographystyle{elsarticle-harv}
\bibliography{ref}

\end{document}